# Phase Transition and Network Structure in Realistic SAT Problems

Soumya C. Kambhampati & Thomas Liu
McClintock High School
Tempe, AZ
{budugu2z,thomasliu02}gmail.com

**Abstract:** A fundamental question in Computer Science is understanding when a specific class of problems go from being computationally easy to hard. Because of its generality and applications, the problem of Boolean Satisfiability (aka SAT) is often used as a vehicle for investigating this question. A signal result from these studies is that the hardness of SAT problems exhibits a dramatic easy-to-hard phase transition with respect to the problem constrainedness. Past studies have however focused mostly on SAT instances generated using uniform random distributions, where all constraints are independently generated, and the problem variables are all considered of equal importance. These assumptions are unfortunately not satisfied by most real problems. Our project aims for a deeper understanding of hardness of SAT problems that arise in practice. We study two key questions: (i) How does easy-to-hard transition change with more realistic distributions that capture neighborhood sensitivity and rich-get-richer aspects of real problems and (ii) Can these changes be explained in terms of the network properties (such as node centrality and small-worldness) of the clausal networks of the SAT problems. Our results, based on extensive empirical studies and network analyses, provide important structural and computational insights into realistic SAT problems. Our extensive empirical studies show that SAT instances from realistic distributions do exhibit phase transition, but the transition occurs sooner (at lower values of constrainedness) than the instances from uniform random distribution. We show that this behavior can be explained in terms of their clausal network properties such as eigenvector centrality and small-worldness (measured indirectly in terms of the clustering coefficients and average node distance).

# Phase Transition and Network Structure in Realistic SAT Problems

## Introduction

A fundamental question in Computer Science is about understanding the computational hardness of problems. Most work on computational hardness characterizes "worst case" complexity of problems—that is, the amount of computational time taken to solve the problem in the worst case. Worst case complexity is, however, not always a good indicator of problem hardness. A class of problems may be hard to solve in the worst case, yet be easy to solve in most practical cases. Consequently, researchers are interested in finding *exactly where the problems go from being easy to hard*. Work on this question has been done in the context of the Boolean Satisfiability problem (SAT for short) [1-5].

Given a set of Boolean constraints on a set of propositions (aka variables), the SAT problem is to find a True/False assignment to the variables so that all the constraints are satisfied. Because SAT models the problem of choice making under constraints, many real-world problems---ranging from scheduling tasks, to circuit verification and planning for robots---can be reduced to SAT. SAT solvers have thus become an integral part in many computational tasks. Consequently, developing an understanding of the difficulty of SAT problems in practice is very important.

SAT problems take exponential time (in the number of variables) to solve in the worst case, but are, however, easy to solve for most cases in practice. Research in the early 90's [1] showed that k-SAT problems (SAT problems with a fixed number of variables, *k*, in every clause) are easy to solve both when the "constrainedness" (the ratio of the number of clauses to the number of variables) is low and when it is high, abruptly transitioning from easy to hard in a very narrow region of constrainedness. Most of this "phase transition" studies were done on SAT instances that follow uniform random distribution. In such a distribution, variables take part in clauses with uniform probability, and clauses are independent (uncorrelated).

The assumptions of uniform random distribution are, however, not satisfied when we consider SAT instances that result from the compilation of real problems. Our project thus aims for a deeper understanding of the hardness of SAT problems that arise in practice. In particular, we study these **two key questions**: *(1) How does the phase transition behavior change with more realistic and natural distributions of SAT problems?* and *(2) Can we gain an understanding of the phase transition in terms of the network structure of these SAT problems?* Our ***hypothesis*** is that the network properties help predict and explain how the easy-to-hard problem transition for realistic SAT problems differs from those for uniform random distribution.

In aid of the first question, we considered SAT instances generated according to two realistic distributions: (i) ***rich-get-richer*** and (ii) ***neighborhood-sensitive***. We developed our own SAT solver and conducted a large-scale systematic empirical study on the phase transition behavior of SAT instances from these distributions. For the second question, we considered the properties of the "*clausal networks*" of the SAT instances and analyzed their properties such as node centrality and average clustering coefficient. Our results show that realistic SAT instances *do exhibit phase transition*, but their transition points differ significantly from the uniform random ones. We shall demonstrate that these differences can be explained in terms of the properties of their network structure.

# Background on SAT & Network Analysis

A Boolean Satisfiability Problem (SAT) is the problem of determining if the variables in a Boolean formula can be set to True/False values so as to make the entire Boolean formula return true. If this is possible, the problem is called satisfiable. If not, the problem is unsatisfiable. **Figure 1** shows a simple satisfiability problem with three variables (*Joe*, *Tom* and Moe) and three constraints (C1, C2 and C3). It can be interpreted as the problem of throwing a party with three of your friends, with the constraints specifying invitation rules. Given *n* variables, there are $2^n$ possible assignments, and in the worst case, a solver has to check through each one of them. Indeed, SAT problems are known to be NP-Complete [1].

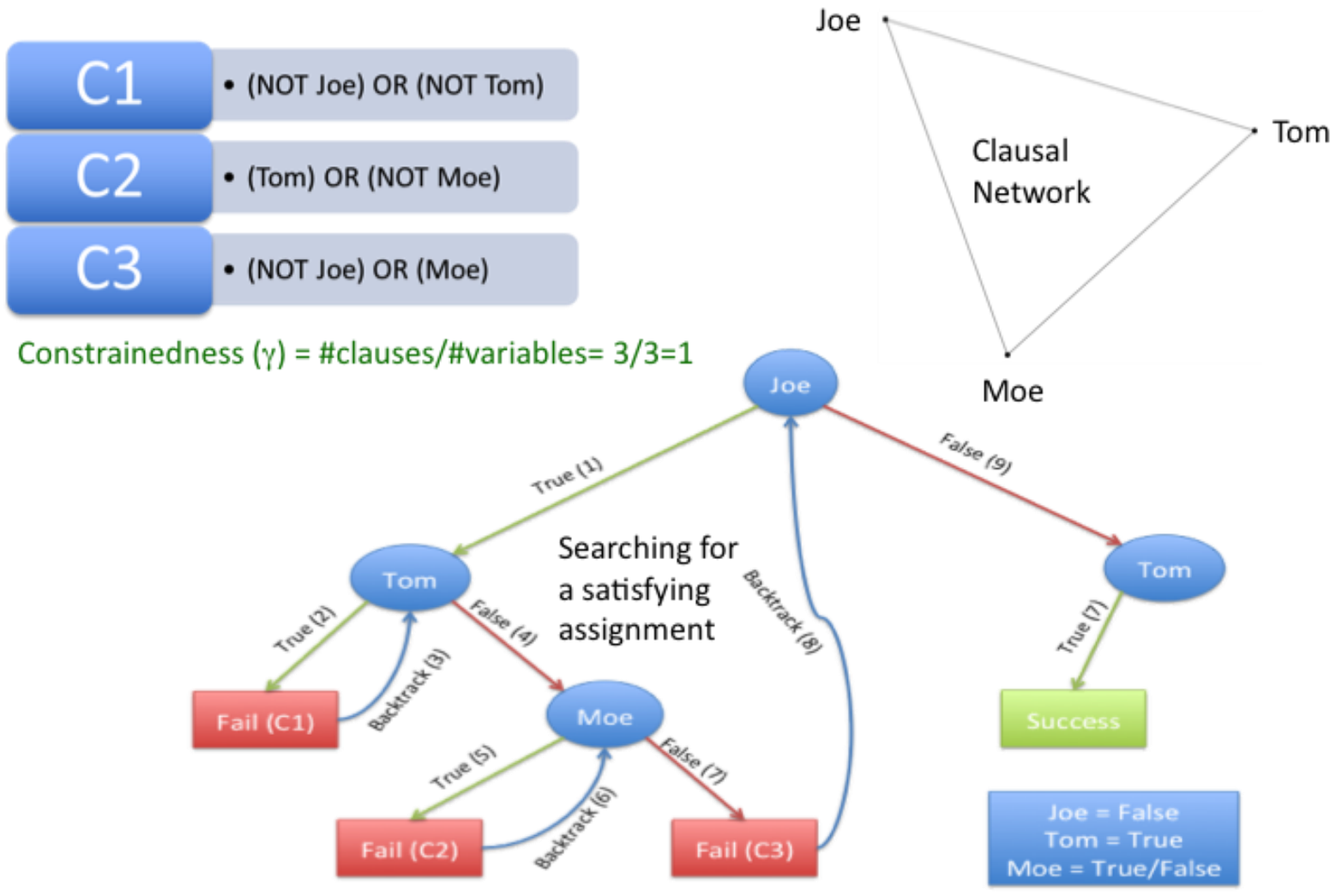

**Figure 1 An example SAT problem and its clausal network.**

**CNF Clauses:** The constraints in a general Boolean satisfiability problem can be any arbitrary propositional logic formula (involving negation, conjunction, disjunction and implication). CNF, or Conjunctive Normal Form, is a simpler (normal) form where each constraint (also called "clause") is a disjunction of normal or negated variables (aka "literals"). Any Boolean formula can be easily converted to an equivalent set of CNF clauses [3]. For example, the constraint (P IMPLIES Q) can be converted to the clause (~P OR Q) (where "~"stands for negation). The length of the clause is the number of variables in it.

**k-SAT:** A general SAT problem, converted to CNF form, might normally result in clauses of varying lengths. A **k**-SAT problem instance is one where all the clauses are exactly length **k**. Interestingly, while 1-SAT and 2-SAT are both easy to solve (worst-case polynomial), from 3-SAT onwards, the worst case complexity becomes exponential. Thus scientists often study 3-SAT problems as a stand-in for the general Boolean satisfiability problems [1,2,4].

**Phase Transition in SAT:** While SAT problems have an exponential worst case complexity, it has long been known that most random SAT instances are easy to solve. This brings up the question: *where are the hard SAT problems?* Researchers have tried to answer this question by experimenting with randomly generated SAT instances [1,2]. They have found that SAT problem instances can be characterized by their *constrainedness* ($\gamma$) which is the ratio of the number of clauses ('constraints') to the number of variables. For the example problem in **Figure 1**, the constrainedness is 1. "Hard problems", or problems that take SAT solvers the maximum effort to solve, occur in a narrow band of constrainedness, with easy problems to the left and right of them (see Figure 2 and Figure 3 for example). In this narrow band, the expected number of satisfying assignments transitions from greater than 1 to less than 1 [1,3,4]. Much of this empirical research has however been focused on uniform random distributions—where each variable takes part in a clause with the same probability. Our aim is to understand the effect of different problem distributions on SAT, both by analyzing the location of the phase transition and by studying the underlying structure of the problem.

**Clausal Network Structure**: We also aim to analyze the SAT problems in terms of their network structure. For this purpose, we define the notion of "*clausal network*" of a SAT problem. This is a network where the nodes correspond to the variables in the SAT problem, and there is a link between two nodes, if the corresponding variables are present together in a SAT clause. **Figure 1** shows the three node clausal network for our example. We intend to analyze these networks in terms of properties such as *clustering coefficient, average path length, proximity ratio and eigenvector centrality* [7,8].

# Materials & Methods

In our quest to find how phase transition behavior and network structure changes with more realistic and natural distributions of SAT problems, we studied two distributions: *rich-get-richer distribution*, and a distribution with *local/neighborhood structure*. Compared to uniform random distributions, the former biases clauses towards variables that are already selected often, while the latter biases clauses so as to be made of variables from the same bucket/neighborhood. We empirically studied the computational hardness of random SAT problems following these two kinds of distributions, with specific attention to their phase transition behavior. We also analyzed the clausal graphs of the various SAT instances in terms of their network properties to understand how they correlate with problem hardness.

Our approach consisted of writing an efficient backtracking SAT solver with unit propagation in Python and using it to solve a large number of random SAT problems with specified degree of *constrainedness*. We instrumented the program to keep track of (a) the amount of CPU time, (b) the number of backtracks and (c) the satisfiability of the instance. We tasked our SAT solver to solve instances generated by neighborhood, rich-get-richer, and uniform random distributions. To ensure *statistical significance*, in all our plots, we considered median values for run time and number of backtracks, and percentage of problems solved over a large sample of SAT instances.  For analyzing the network properties, we used the NetworkX library [9].

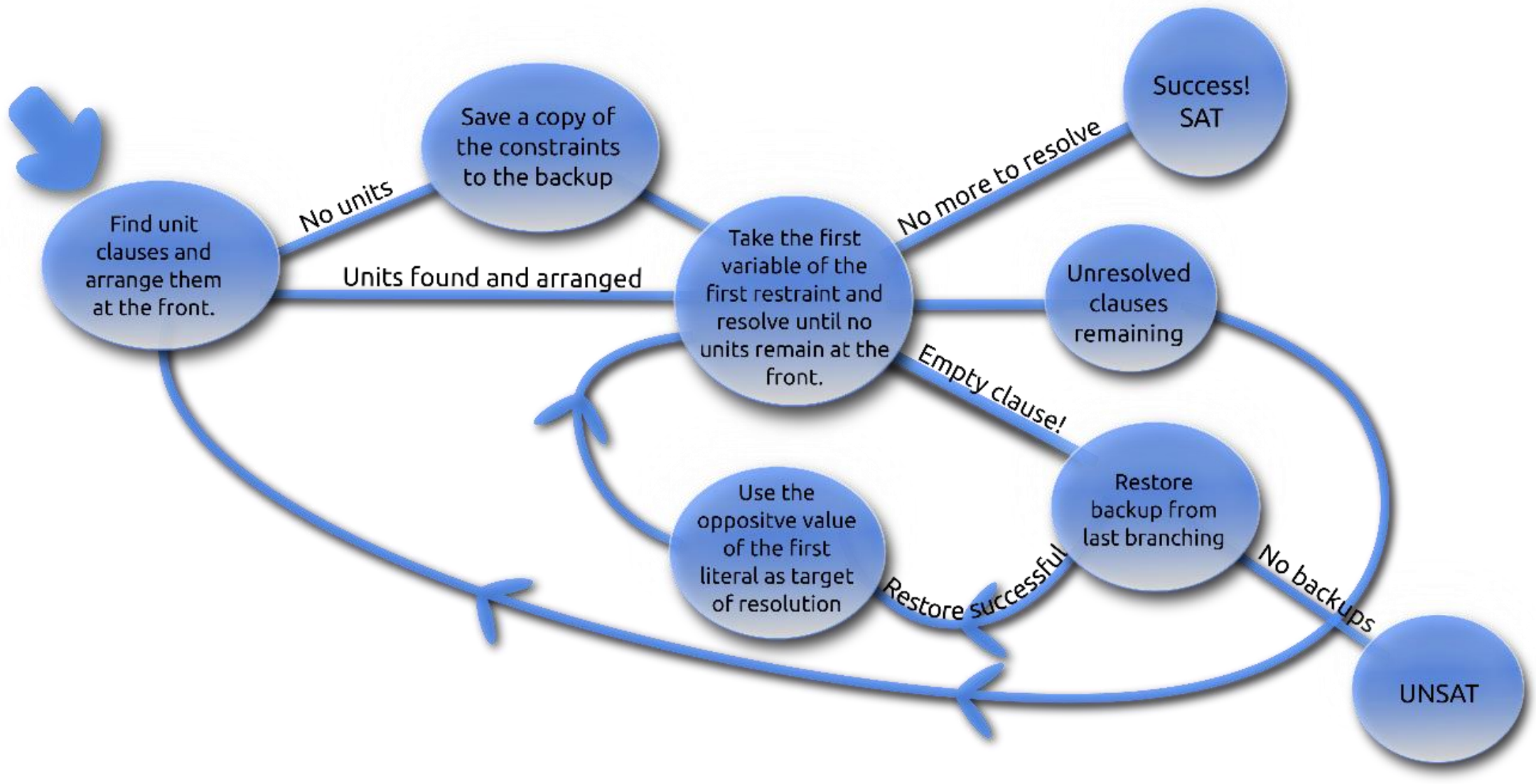

**Figure 2. Flowchart of our SAT Solver algorithm with Unit Propagation**

SAT Solver: **Figure 2** shows the flowchart of our SAT solver. The solver works with CNF clauses in the form of a Python list. The main CNF list contains smaller lists, each of which represents a constraint. Each constraint has a number of literals, represented by integers, and is terminated by ‘0’  (for technical reasons explained below). The program begins by searching for unit clauses (lists that only have two entries, a literal and the ending 0). If it finds any, it puts them all at the beginning of the list. If not, it selects the first variable of the first clause to branch on (with only one choice possible if the first clauses is a unit clause) and saves the whole CNF to memory as a backup. Every time a literal is added to the solution, a list with only the literal (excluding the trailing 0) is put at the end of the current copy of the CNF. In this way, during *unit propagation*, a variable that was propagated before is not propagated again. It then resolves the constraints by taking the first variable of the first constraint and resolving it with the rest until it has to branch again or the solution is obtained, or an empty clause occurs. Resolution is accomplished by removing entire clauses with the literal inside, and by removing the negation of the current literal that is the target of resolution from clauses. If resolution produces an empty clause, it backtracks to the last branching (by restoring the appropriate copy of the entire CNF previously saved) and branches on the opposite literal instead.

**Uniform Random Distribution:** Uniform Random Distribution serves as a control for our study, as prior research [1,2] already establishes where the phase transition for this case exists. SAT instances under this distribution have clauses that are formed by picking variables with uniform probability. To generate the problem instances, we wrote a program that takes **v,** the number of variables in the SAT instance, and the constrainedness parameter $\gamma$ and makes **v*** $\gamma$ random clauses. Each clause (constraint) consists of three randomly chosen variables, and each variable is then randomly chosen to occur as a positive or negative literal. We first focused on 3-SAT (all clauses of size 3), and subsequently also experimented with 2-SAT and 4-SAT problems.

We generated 100 SAT problem instances for every increment of 0.1 of $\gamma$ from 0 to 10, and gave each to our SAT solver. We graphed the data on problem hardness and percentage satisfiable instances at each level of constrainedness ($\gamma$). The hardness is measured in terms of median CPU time and the number of backtracks.

**Rich get Richer Distribution:** Rich-get-richer is a kind of distribution in which once a variable is chosen to take part in a clause, the probability that it will be picked again is increased. Thus "rich" variables which got picked once, get picked more. We explored this distribution since in many real world problems, the rich do indeed get richer: there are certain "critical variables" that wind up being part of many problem constraints, and clauses tend to be correlated.

We generated the SAT instances for this distribution by writing a program that initially creates a list of all of the variables that would be used in the problem. Then, one variable is randomly chosen to be the first literal, and a copy of it is inserted into the list (so that two instances of the variable now exist). Note that the variables with more instances have a higher chance of being picked once again. Our SAT generator program loops through this method creating a number of clauses equal to $\gamma$ times the number of variables. The resulting SAT instances did indeed have certain variables appearing in clauses much more often than others, as expected. We then took these randomly-generated SAT problems and tasked our SAT solver to solve them, running 100 samples for each 0.1 increment of $\gamma$ (from 0 to 10). Finally, we graphed the output and observed if phase transition occurred, and if so, where it occurred.

**Neighborhood-sensitive Distribution:** Neighborhood distribution is a distribution with local structure where variables are initially placed into "buckets". To generate a clause, we start by picking a variable from a bucket **b.** Each successive variable for this clause is selected from any of the buckets with probability ***p***, and from the same bucket **b** with a probability *1 – **p***. We explored this distribution because in many real problems, there are loosely coupled clusters of variables such that there are more constraints among the variables of the same cluster than across different clusters. The parameter ***p*** thus measures the degree of globalness (i.e., not neighborhood sensitive) of the distribution. If ***p****=1* then each variable can come from any of the buckets (just like uniform random distribution), while if ***p****=0,* all of the variables constituting a clause come from the same bucket—thus each cluster is disconnected (in terms of constraints) from other clusters.

We implemented this distribution by writing a program in Python that accepts **p** and generates clauses as described above. To ensure that we had implemented neighborhood distribution correctly, we checked if a problem with the probability ***p****=1* mirrored the empirically shown graph of uniform random distribution. We varied the probability ***p*** from 0 to 1.0 in increments of 0.3. We generated 100 problem instances for every increment of 0.1 for $\gamma$, and graphed the median number of backtracks for each.

**Sudoku Instances:** Although our main focus is on random SAT instances, in the course of our project, we also experimented with SAT instances derived from Sudoku problems. We obtained the initial SAT constraints for the Sudoku rules from an online converter [10,11]. Using this method, we converted the given Sudoku values into CNF using a Python program we wrote.

**Clausal Network Structure Analysis:** In order to better understand the properties of SAT problems and the effect of different problem distributions on them, we converted SAT problems into networks so that we could use social networking analysis (SNA) techniques to study the underlying structure of the problems. We compiled SAT problems into networks by creating a *node* for every variable in a SAT problem, and creating a *link* between two nodes if two variables were found in the same clause. We then compared the clausal networks of the three distributions (rich-get-richer, neighborhood, and uniform random) through four metrics [7,8]: (i) the **eigenvector centrality** of the nodes, which measures the degree of centrality of the nodes (and thus their relative influence) in the network, (ii) the **average clustering coefficient** of the nodes, or how clustered the nodes in the network are, (iii) the **average length**, or the average shortest distance between any two nodes (note that the distance is measured by the number of links one must traverse to get from the starting node to the end node), and (iv) the **proximity ratio**, a metric for measuring the small-world characteristics of a network [6,7]. We used the library NetworkX [9] to extend Python so that it could work with clausal networks.

# Results & Analysis

In the following, we will first present the results from the clausal network analysis, and then the results from phase transition experiments. Our analysis will also demonstrate how the network properties help explain the phase transition results.

## Findings on the Clausal Network Structure of SAT Problems

Recall that we convert a SAT instance into a clausal network by having nodes correspond to SAT variables, and placing a link between two nodes if they take part in a SAT clause together. In the following, we present and analyze the properties of the clausal networks of the various distributions to gain insight into their structure.

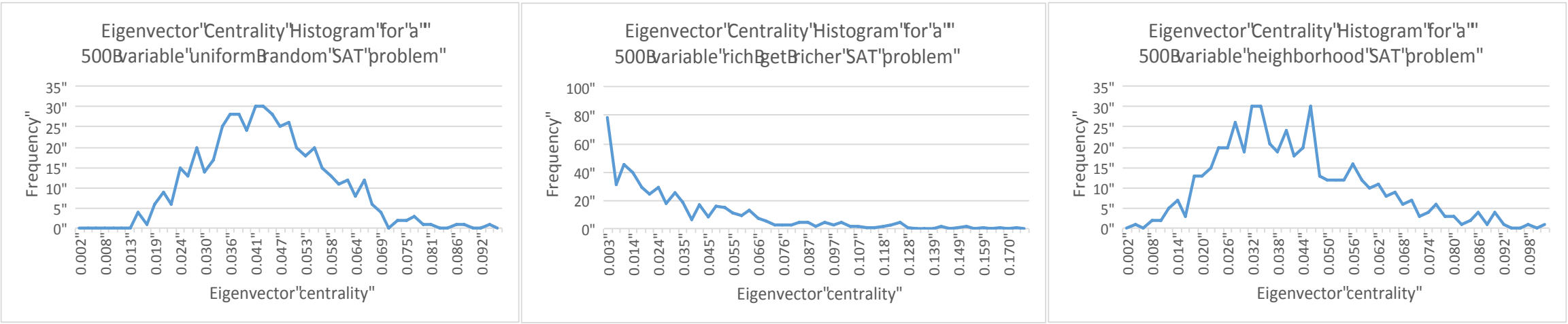

**Figure 3: Frequency histograms of the eigenvector centralities of all variables in 500-variable SAT instances (with $\gamma$ = 4.3) drawn from uniform random (left), rich-get-richer (center) and neighborhood (right).**

**Eigenvector Centrality:** We started by analyzing the centrality of the nodes in the clausal networks for each of the distributions. We experimented with 500 variable SAT instances, computed the eigenvector centrality measure for the nodes, and generated histograms of the node centrality. **Figure 3** shows the results for the three distributions. For the uniform random distribution (plot on the left), we see that most of the centrality values are near the median with only a few slightly more or slightly less. This indicates that for the uniform random distribution, most nodes, or variables, are generally of equal influence, with the exceptions being of very low probability. On the other hand, the histogram for the rich-get-richer distribution (middle plot) displays a distinct long tail, with a large number of non-influential nodes and a long tail of increasingly influential ones. The neighborhood distribution (plot on the right) straddles the two, with more influential nodes than those in uniform, but less than those in rich-get-richer. In the rich-get-richer distribution, a few key variables are present in a large number of clauses, making them vastly more influential compared to the rest of the variables. Solving these instances should thus be easy once the influential variables are assigned values.

Our next set of experiments consisted of taking twenty 50-variable 3SAT problems for each 0.1 increment of $\gamma$ from 1 to 10 and calculating the average clustering coefficient, the average length, and proximity ratio of each; **Figure 4** shows the results.

**Average Clustering Coefficient**: The clustering coefficient is a measure of the degree of connectedness of a node's neighbors. **Figure 4** (left plot) shows that the average clustering coefficient increases with $\gamma$ for each of the three distributions. This should be expected since increasing $\gamma$ increases the number of constraints (and thus links in the clausal graph), eventually leading to a fully clustered network. The more interesting observation is that the *rate of increase* is higher for the rich-get-richer and neighborhood distributions. Moreover, for any given value of $\gamma$, we see that the rich-get-richer and neighborhood distributions produce SAT instances with more clustered clausal networks.

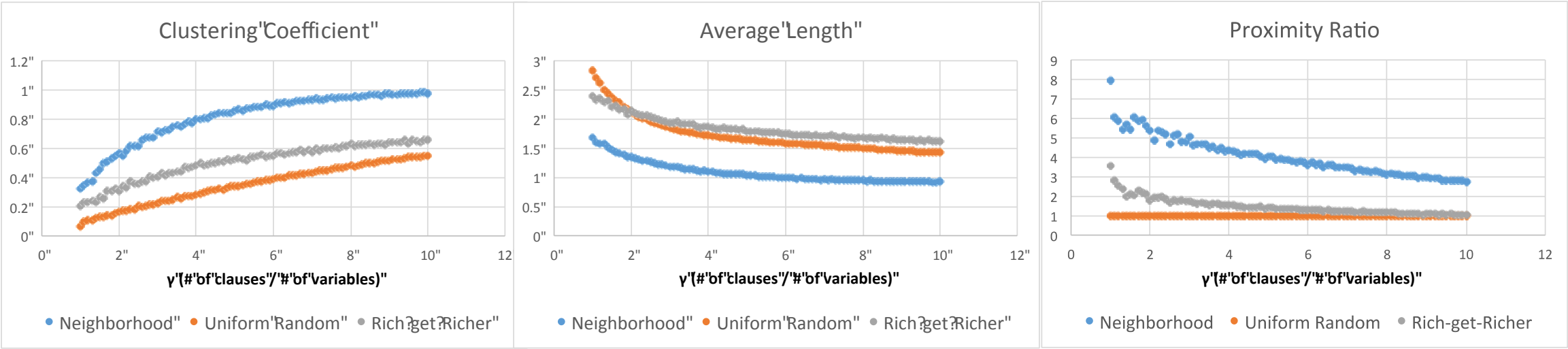

**Figure 4: Plots of average clustering coefficient, average length, and proximity ratio in 50 variable SAT instances with uniform random (orange), rich-get-richer (gray) and neighborhood (blue) distributions**

**Average Length**: The middle plot in **Figure 4** shows the average length, which is how far two random connected nodes are from each other in the network. Once again, the overall trend of decrease of average lengths with increasing $\gamma$ can be explained in terms of the increase in the number of constraints (and thus links). As links are added, average length decreases because on average, it becomes easier to traverse from one node to another. What is interesting is that the average length at any $\gamma$ is significantly lower for the neighborhood distribution. Since each variable is within its own neighborhood, and neighbors become more connected within these neighborhoods, it becomes increasingly easy to traverse from node to node inside a neighborhood.

**Proximity Ratio**: The right plot in **Figure 4** shows the proximity ratio, which is a measure proposed by Walsh [7] to assess the "small worldliness" [6] of a network. It compares the ratio of how clustered a network is to its average length with the ratio of a corresponding uniform random network. Thus, by definition, the uniform random distribution has a proximity ratio of 1. The neighborhood distribution, in contrast, shows a very high proximity ratio because it has a really high clustering coefficient and a very low average length. Thus, it conforms to the idea of a "small world" network [6] as several loosely connected clusters. Rich-get-richer distribution also has a higher proximity ratio (especially for low γ values), although it is not as pronounced as the neighborhood one.

We also analyzed the clausal network properties of the SAT instance corresponding to a Sudoku problem. We found that like neighborhood and rich-get-richer distributions, Sudoku instance also has a very high proximity ratio (of 13.74, derived from an average clustering coefficient of 0.291 and an average length of 2.55) showing its pronounced small world nature [7].

# Findings on the Phase Transition

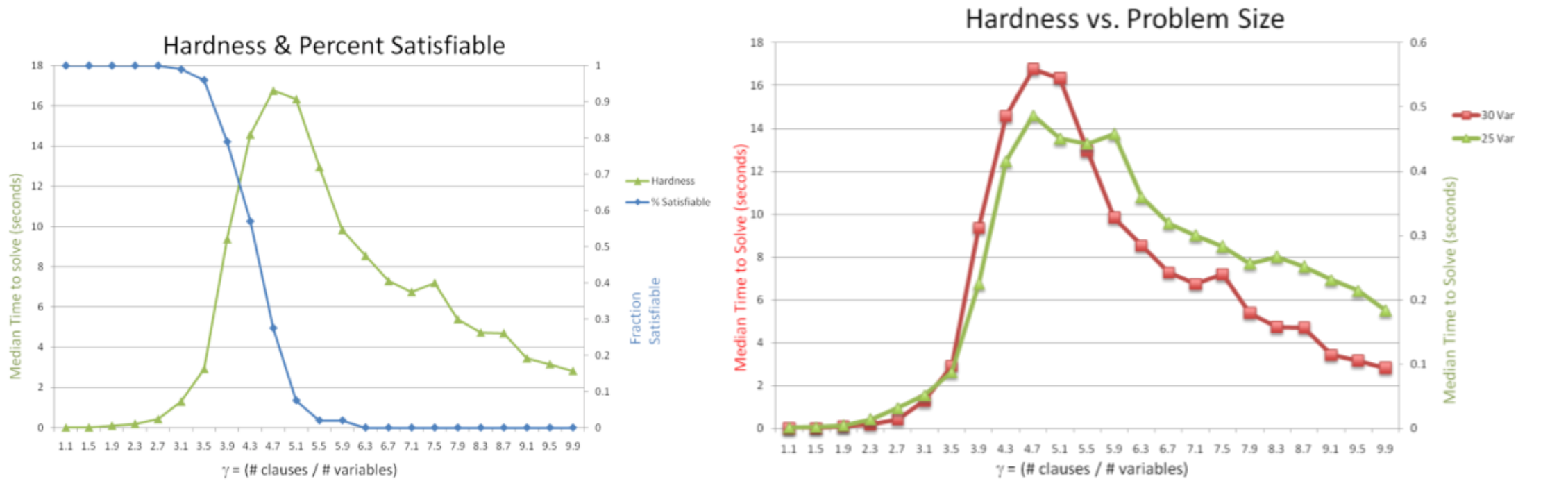

**Figure 5: Phase transition of 3-SAT problems under uniform random distribution. Notice that the transition in hardness occurs approximately where 50% of the instances are solvable The hardness phase transition occurs at the same place, but is more pronounced as the size increases.**

**Uniform Random Distribution**: The results for uniform random distribution are shown in **Figure 5**. For 3SAT, we noted that the phase transition occurs around γ=4.3, where average hardness peaks, and the percentage satisfiable reaches the half-way (50%) point. We also note (from Figure 3) that the transition is more prominent as the size of the SAT instances (as measured in number of variables) increases. These are in accordance with results from prior research [1,2,4].

We also experimented with 2-SAT and 4-SAT instances (see **Figure 6**) and found that the transition for 2-SAT occurs at γ=1, and for 4-SAT it occurs around γ=9.76, We did analyze why the transition point increases as *k* increases in *k-SAT.* As *k* increases, the clause length increases, and so the clauses become "looser" and thus easier to satisfy (recall that CNF clauses are a disjunction of literals). Thus you need more loose clauses to constrain the problem to the same degree as the tight clauses. This explains why the phase transition point increases as *k* increases. For the rest of our analysis, we restricted our attention to 3-SAT problems.

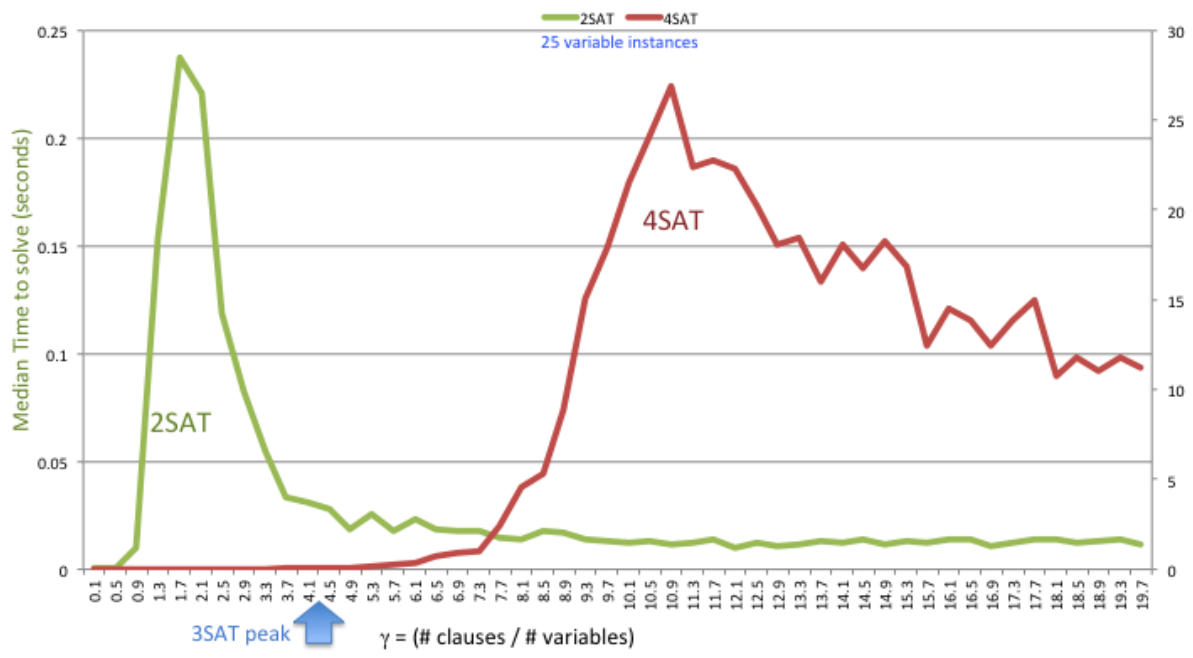

**Figure 6. Comparing phase transition of k-SAT problems as clause length (k) increases. Notice that the phase transition point shifts right (i.e., occurs at higher γ values) as k increases**

**Rich-get-Richer Distribution:** We tested the rich-get-richer distribution with SAT instances with 150 variables. The results are shown in **Figure 7**. We found that the distribution *does exhibit phase transition*, but the transition occurs at around γ=2.8, a value significantly less than 4.3, the phase transition point for uniform random distribution. This can be *explained* in terms of its network properties (**Figures 3** and **4).**

Specifically, as shown by the long-tailed node centrality in **Figure 3**, clauses in the rich-get-richer distribution are centered around a smaller set of variables.  Satisfiability drops earlier because the high concentration of clauses surrounding a certain variable raises the chances of unsatisfiability. To test this hypothesis, instead of adding one copy of the chosen variable back into the selection pool, we experimented with adding an increasingly larger number of copies.  As shown by the right plot in **Figure 7**, not only did this make the transition occur at an even lower  $\gamma$ value, but it also decreased the amount of time required to solve the problem. This is explained by the fact that the SAT solver actively looks for unsatisfiable clauses and declares the entire problem unsatisfiable if one variable has forced contradictory values. Additionally, because of the centralized nature of the rich-get-richer distribution, satisfiable problems are also solved faster: after the few key variables around which the problem is centered are determined, the rest fall into place easily through unit propagation.

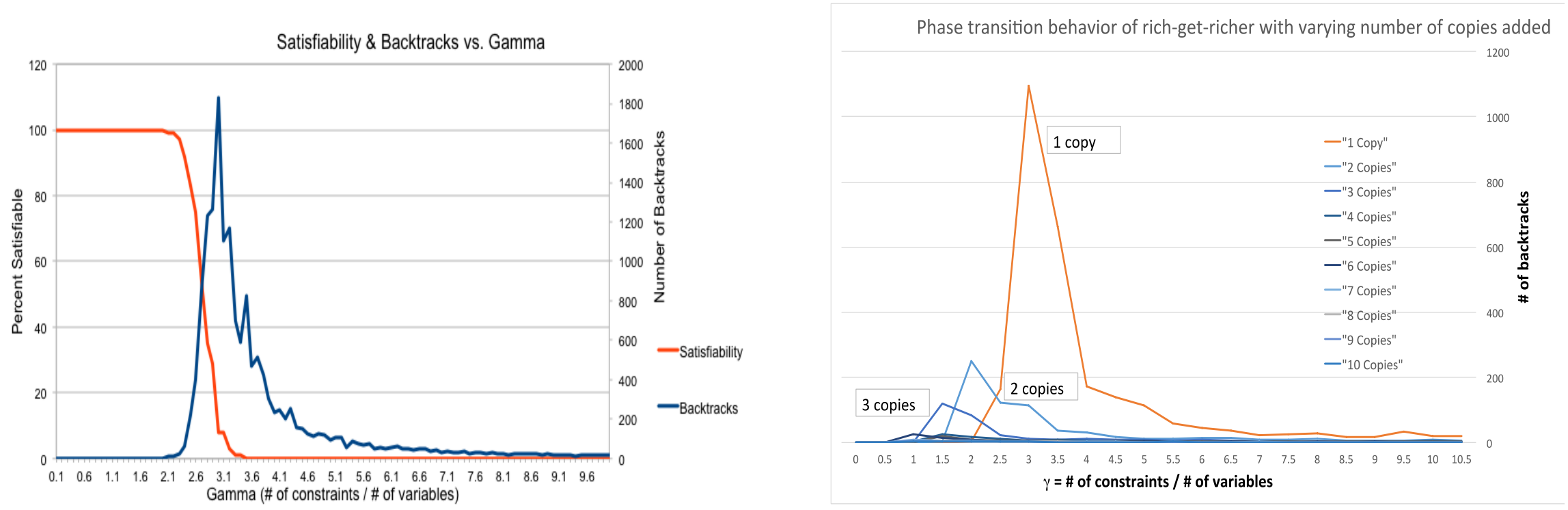

**Figure 7. Phase transition behavior under rich-get-richer distribution for 3-SAT.  The left plot shows that the transition occurs at a lower γ value compared to uniform random distribution. The right plot shows that as we increase the "degree of richness" (that is, the number of copied added), the transition point shifts further down, and the hardness of the problems falls.**

**Neighborhood-sensitive Distribution:** The results for the neighborhood distribution are shown in **Figure 8**. With the neighborhood distribution,  once again, we found that the phase transition *does occur*, but  it shifts depending on the neighborhood parameter, **p**. Recall that **p** determines how likely the literals of a clause are going to be chosen from any bucket (neighborhood) rather than remaining within the initial bucket. As **p** approaches 1, the problem becomes identical to a uniform random distribution problem. However, when **p** is close to zero, the variables of each clause increasingly belong to the same bucket, effectively rendering the entire SAT problem a union of multiple smaller, *disconnected* SAT problems.

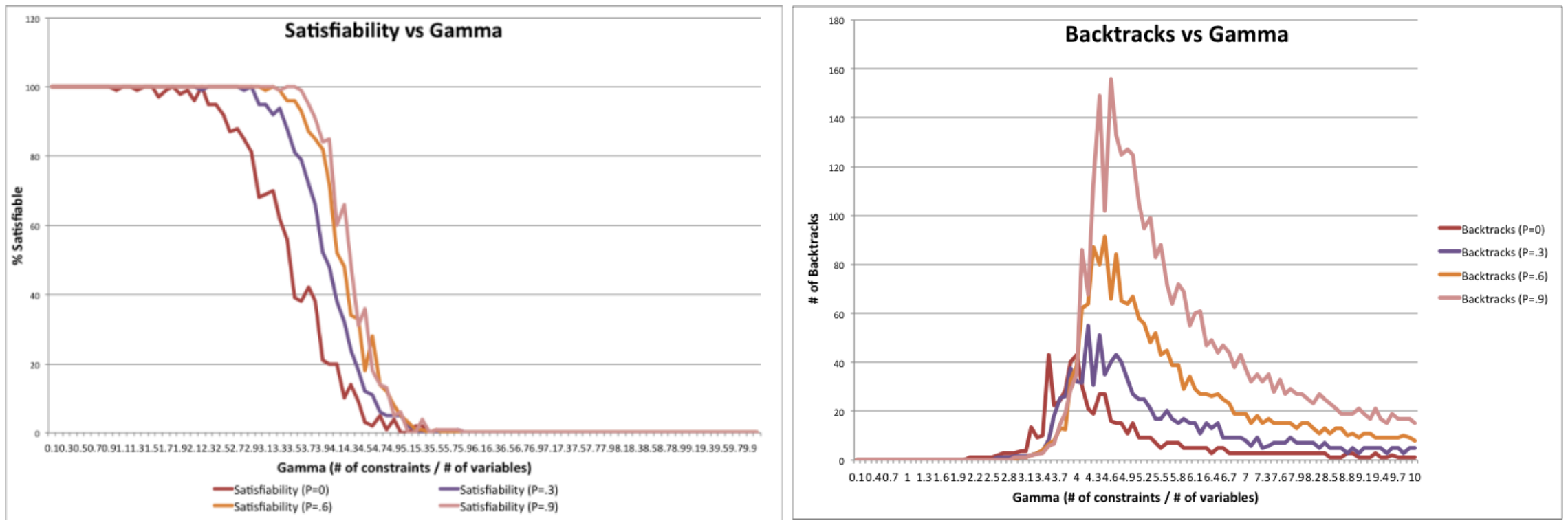

**Figure 8. Percentage satisfiable problems , and average hardness of 3-SAT as the neighborhood parameter p is varied. Note that p=1 corresponds to uniform random SAT. We see that phase transition occurs earlier if the problems are more sensitive to the neighborhood. Note that p=1 corresponds to uniform random distribution. We see that although phase transition occurs earlier for p<1, the difficulty of solving the instances is lower, because we have loosely coupled clusters of fewer variables.**

As **p** is varied from 1 to 0, we note (right plot) that he phase transition occurs earlier, and the median number of backtracks required to solve the problem drops significantly. This can be explained by the fact that as **p** goes to zero, the SAT problem gets increasingly localized into several smaller, distinct problems. These observations are also confirmed (**Figure 4)** by the high average clustering coefficient, the characteristically small path length, and its high proximity ratio. These network measures indicate the small world characteristic of neighborhood distribution, where several groups of variables cluster together with heavy connections among each other. The rate of satisfiability (left plot) also drops earlier with smaller problems. This is because with several smaller problems, only one local problem has to be unsatisfiable for the entire problem to be unsatisfiable; the chain is only as strong as its weakest link. Furthermore, each smaller problem, with a correspondingly smaller neighborhood of variables to choose from, is more constrained, and thus requires fewer clauses to achieve the same amount of constrainedness.

# Conclusion & Future Work

Prior research has shown that random SAT problems exhibit a dramatic easy-to-hard transition in problem hardness as their constrainedness  is varied. The main focus of that past work has, however, been on uniform random distributions, in which clauses are generated independently of other clauses.  Our project aimed for a deeper understanding of  hardness of realistic SAT problems that arise *in practice*. We studied two key questions: (i) *How does the easy-to-hard transition change with more realistic distributions that capture neighborhood sensitivity and rich-get-richer aspects of real problems* and (ii) *Can these changes be explained in terms of the clausal networks of the SAT problems*.  We experimented with SAT instances from rich-get-richer and neighborhood-sensitive distributions, both of which make more realistic assumptions than uniform random distributions.

Our analysis of the clausal networks of rich-get-richer and neighborhood distributions established that they both have very different network properties than problems following uniform random distributions. We also found that these distributions *do exhibit phase transition*, but the relative locations of their transition, as compared to those of uniform random distribution, shift. Our work determined that the phase transition for the neighborhood distribution occurred earlier, the more localized the clauses were, and that the phase transition for the rich get richer distribution also occurred earlier than that of the uniform random distribution. These distributions also require fewer backtracks to solve. We also showed that the shift in phase transition can be explained in terms of their clausal network properties such as eigenvector centrality and small-worldness (measured indirectly in terms of the proximity ratio, clustering coefficients and average node distance). Localized clauses create smaller problems that have higher constrainedness and cause the problem to fail earlier, and the centralization of the rich-get-richer around a small set of variables (as shown by our analysis of eigenvector centrality) also decreases the satisfiability and hardness of the problem.

Understanding when specific computational tasks become hard is a core challenge in Computer Science. SAT problems are especially attractive as many other real world problems involving decision making under constraints, such as planning, scheduling, and circuit verification, can be translated to SAT. We believe that the results of our project can have significant implications on the understanding of computational hardness in real problems, by shedding light on their phase transition and clausal network properties. In the future, we hope to realize this potential by (i) working towards analytically characterizing the phase transition points in terms of the distribution parameters and (ii) predicting the computational hardness of industrial decision making problems, such as scheduling and circuit verification, in light of our results from realistic SAT distributions.